%% file: main.tex
\documentclass{article}

\usepackage[utf8]{inputenc} 
\usepackage[T1]{fontenc}    
\usepackage{hyperref}       
\usepackage{url}            
\usepackage{booktabs}       
\usepackage{amsfonts}       
\usepackage{nicefrac}       
\usepackage{microtype}      
\usepackage[table]{xcolor}         
\usepackage{graphicx}
\usepackage{subcaption}
\usepackage{caption}
\usepackage{hyphenat}
\usepackage{setspace}
\usepackage{amsmath}
\usepackage{amssymb}
\usepackage{mathtools}
\usepackage{amsthm}
\usepackage{longtable}
\usepackage{multirow}
\usepackage{float}
\usepackage{tabularx}
\usepackage{enumitem}
\usepackage{threeparttable}
\usepackage{algorithm}
\usepackage{algorithmic}
\usepackage{titletoc}
\usepackage{pifont}
\usepackage{wrapfig}

\newcolumntype{C}{>{\centering\arraybackslash}X}

\newcommand{\cmark}{\textcolor{teal!80!black}{\ding{51}}}   
\newcommand{\xmark}{\textcolor{red!70!black}{\ding{55}}}     

\definecolor{prompttitlebg}{HTML}{75999B} 
\definecolor{promptbodybg}{HTML}{F1F6F7}  

\newcommand{\thinktag}[1]{\textcolor{blue}{\textless #1\textgreater}}
\newcommand{\retrievetag}[1]{\textcolor{teal}{\textless #1\textgreater}}

\newcommand{\actiontag}[1]{\textcolor{red}{\textless #1\textgreater}}
\newcommand{\emptytag}[1]{\textcolor{purple}{\textless #1\textgreater}}

\usepackage[preprint]{my_style}

\usepackage{tcolorbox}
\tcbuselibrary{skins, breakable, listings}

\definecolor{brandblue}{RGB}{57,95,207}
\definecolor{linkblue}{HTML}{0064E0}
\definecolor{textgray}{HTML}{1C2B33}
\definecolor{boxbg}{HTML}{F1F4F7}

\hypersetup{
  colorlinks=true,
  linkcolor=brandblue,
  citecolor=brandblue,
  urlcolor=linkblue
}

\newcommand{\paperTitle}{MemHarness: Memory Is Reconstructed, Not Replayed}

\newcommand{\paperAuthors}{%
  {\sffamily\bfseries Rong Wu$^{1,2}$}, 
  {\sffamily\bfseries Daocheng Fu$^{2,3}$}, 
  {\sffamily\bfseries Licheng Wen$^{2,4,5}$},
  {\sffamily\bfseries Xuemeng Yang$^{2}$},
  {\sffamily\bfseries Shu Zou$^{6}$},
  {\sffamily\bfseries Jianbiao Mei$^{2}$},
  {\sffamily\bfseries Yuxin Wang$^{7}$},
  {\sffamily\bfseries Hairong Zhang$^{2}$},
  {\sffamily\bfseries Yu Yang$^{1,2}$},
  {\sffamily\bfseries Tao Hu$^{2}$},
  {\sffamily\bfseries Cong Zhang$^{1}$},
  {\sffamily\bfseries Botian Shi$^{2}$},
  {\sffamily\bfseries Pinlong Cai$^{\dag,2}$},

}

\newcommand{\paperAffiliations}{%
  {\normalsize $^1$ Zhejiang University}, 
  {\normalsize $^2$ Shanghai Artificial Intelligence Laboratory}
  {\normalsize $^3$ Fudan University}, 
  {\normalsize $^4$ Shanghai Innovation Institute}
  {\normalsize $^5$ Shanghai Jiao Tong University}, 
  {\normalsize $^6$ The Australian National University}
  {\normalsize $^7$ University of Science and Technology of China}
}

\newcommand{\paperNotes}{%
  {\small $^\dag$ Corresponding Author}%
}

\newcommand{\githubLink}{\url{https://github.com/KnowledgeXLab/MemHarness}}
\newcommand{\publishDate}{\today}

\newcommand{\renderFrontBox}{%
    \tcbset{
    enhanced, frame hidden,
    colback=boxbg,
    left=0.5cm, right=0.5cm, top=0.5cm, bottom=0.5cm,
    arc=16pt,
    before skip=0pt,
    grow to left by=1.5pt, grow to right by=1.5pt,
    overlay={
    \node[anchor=north east, at=(frame.north east), xshift=-2.3cm, yshift=-0.5cm] 
        {\includegraphics[height=1cm]{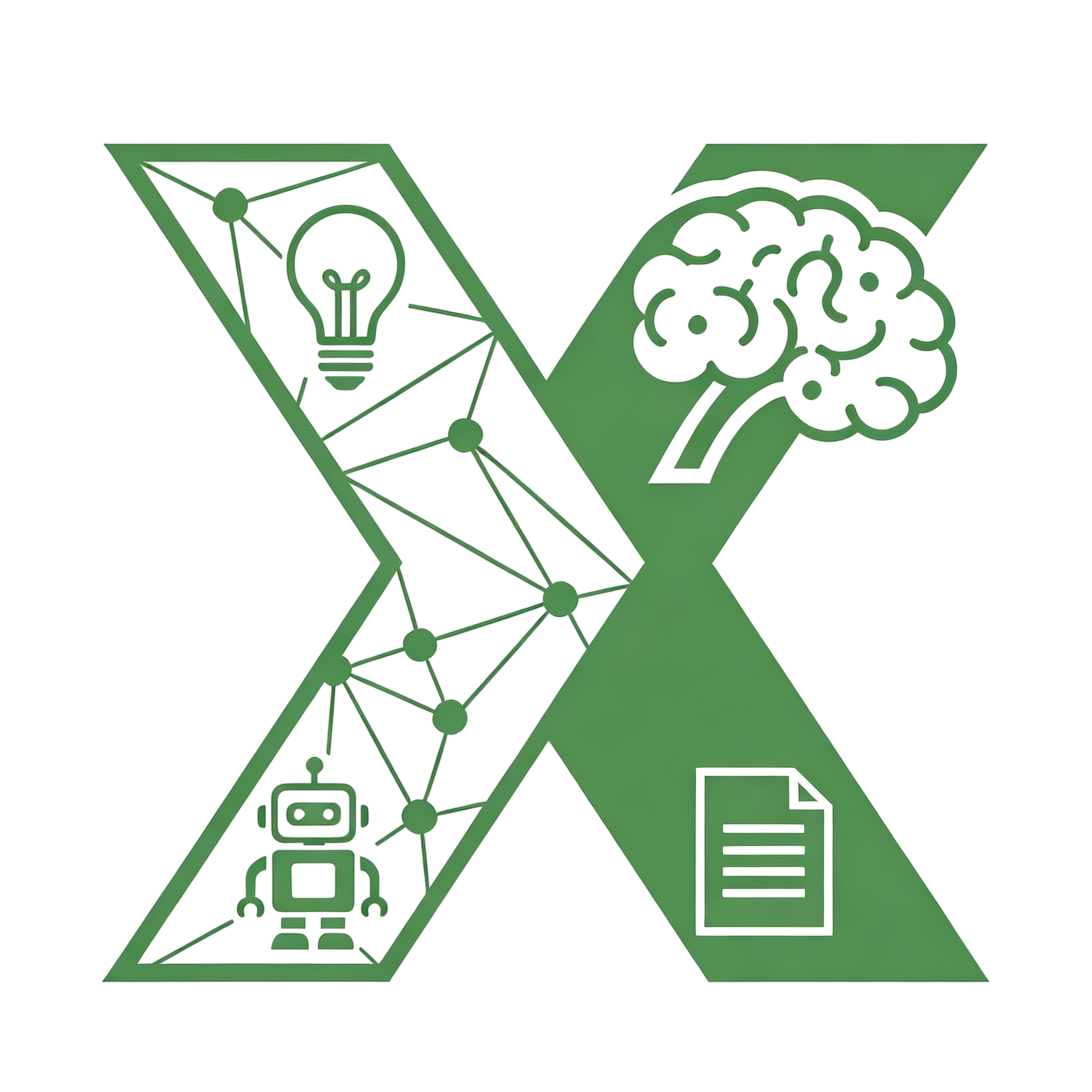}};
    \node[anchor=north east, at=(frame.north east), xshift=-0.5cm, yshift=-0.5cm] 
        {\includegraphics[height=1cm]{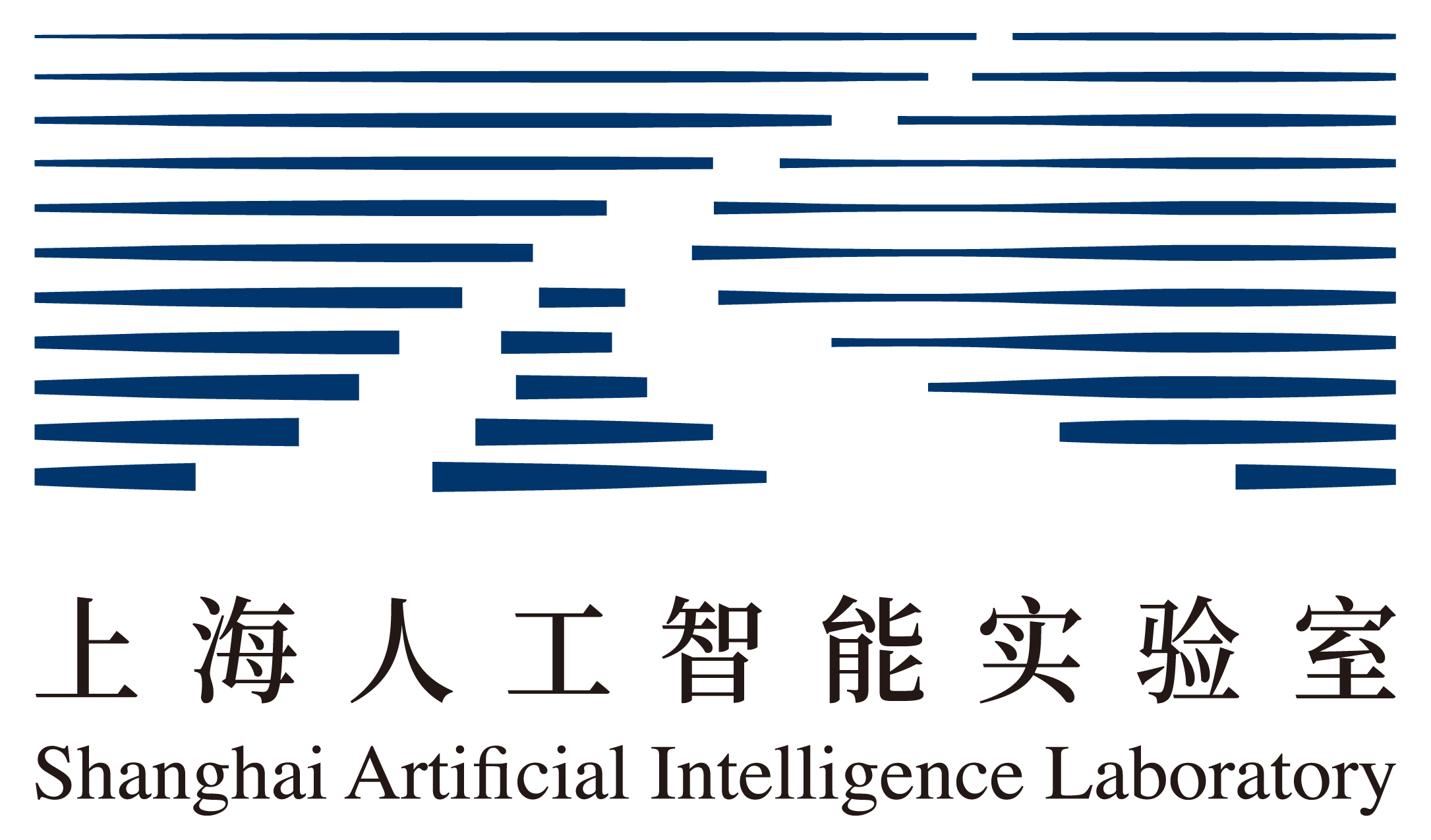}};
    }
  }%
  \begin{tcolorbox}
    \setlength{\parindent}{0cm}
    \setlength{\parskip}{0.5cm}
    {
      \setlength{\parskip}{0cm}
      \raggedright
      \nohyphens
      {
        \vskip 1cm 
        \setstretch{1.4}
        {\huge\sffamily\bfseries\textcolor{black}{\paperTitle}}\par
      }
      \vskip 0.25cm
      \paperAuthors\par
      \vskip 0.35cm
      \paperAffiliations\par
      \vskip 0.08cm
      \paperNotes\par
    }
    \vskip 0.2cm
    {\color{textgray}%
    \input{sections/abstract}
\par}
    \vskip 0.4cm
    {
      \setlength{\parskip}{0cm}
      {\small {\sffamily\bfseries \raisebox{-0.2em}{\includegraphics[width=0.025\linewidth]{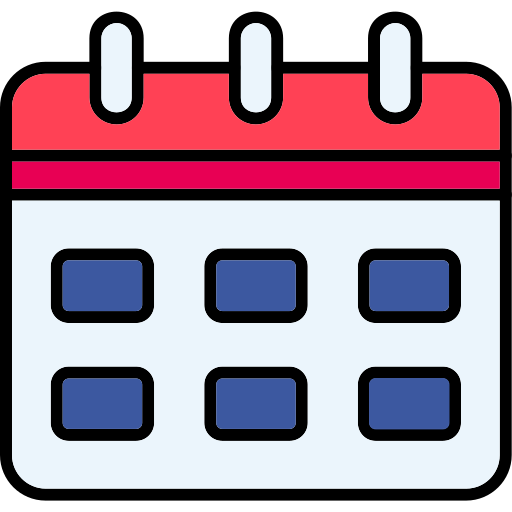}}~~Date:} \publishDate}\par%
      \vskip 0.08cm%
      {\small {\sffamily\bfseries \raisebox{-0.2em}{\includegraphics[width=0.025\linewidth]{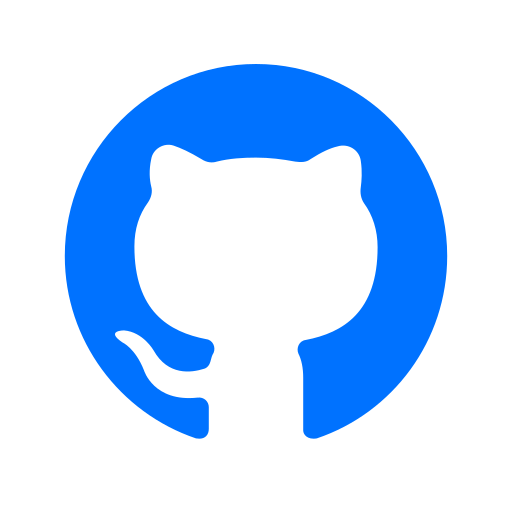}}~~Github Repo:} \githubLink}
    }
  \end{tcolorbox}
  \tcbset{reset}
}

\begin{document}

\newgeometry{top=1in, bottom=0.75in, textwidth=6.3in, textheight=9in}
\renderFrontBox


\input{sections/introduction}

\input{sections/related_work}

\input{sections/methodology}

\input{sections/experiments}

\input{sections/conclusion}

\clearpage
{
\bibliographystyle{unsrt}  
\bibliography{preprint}
}

\clearpage
\newgeometry{
  textheight=9in, textwidth=5.5in, top=1in,
  headheight=12pt, headsep=25pt, footskip=30pt
}

\input{sections/appendix}

\end{document}

%% file: sections/abstract.tex
Retrieving past experiences has become a common strategy to enhance large language model agents. However, most existing memory-augmented agents treat retrieved experiences as static records to be replayed verbatim, injecting them into the context regardless of whether they align with the agent's current situation. This ``replay'' paradigm ignores the gap between the abstract, general nature of stored experience and the concrete, ever-changing states encountered at decision time, frequently causing negative transfer. In contrast, humans rarely recall past experiences verbatim; instead, they reorganize and adapt retrieved memories to fit the present context. Inspired by this, we propose \textbf{MemHarness}, a framework that equips LLM agents to actively harness and reconstruct past experiences based on the present context. At each decision step, a unified policy model critiques and reconstructs the retrieved experience conditioned on the current state, producing context-grounded guidance before acting. This reconstructive ability emerges naturally through end-to-end training with GRPO. Experiments on ALFWorld and WebShop show that MemHarness substantially outperforms pure RL and static memory-augmented baselines, demonstrating strong robustness in out-of-distribution (OOD) scenarios. Furthermore, our analyses reveal that this reconstruction objective not only prevents negative transfer but also serves as latent guidance during training, fundamentally improving the agent's intrinsic reasoning capabilities.

%% file: sections/introduction.tex
\section{Introduction}
\label{sec:intro}

Large Language Models (LLMs) have demonstrated strong capabilities as autonomous agents for sequential decision-making~\cite{chen2023introspective,fu2025re,jin2025searchr1trainingllmsreason,mei20252}. Experiential memory further enables these agents to reuse prior successes and avoid repeated mistakes~\cite{wang2024comprehensivesurveycontinuallearning,yan2025memory,xu2026mem}. However, most memory-augmented agents follow a \textbf{verbatim replay} paradigm: retrieved trajectories or principles are treated as static records and directly inserted into the model context~\cite{yang2026towards,zhang2025gmemorytracinghierarchicalmemory}. This design conflates retrieval relevance with action-level applicability. A memory may be semantically relevant to the task yet inappropriate for the current interaction state, because it was formed under different environmental conditions. An alternative line of work internalizes experience into model parameters, enabling state-conditioned behavior without explicitly replaying retrieved records~\cite{zhang2025memgen}. While such parametric memory offers adaptive generation, the underlying experience and its influence on a decision remain implicit and difficult to inspect or revise. Explicit memory banks provide greater traceability, but typically lack this adaptive use of experience. This exposes a desirable middle ground: retaining explicit, attributable memories while dynamically reconstructing them for the current state.

\begin{wrapfigure}[36]{r}{0.55\textwidth}
    \centering
    \vspace{-1em}
    \includegraphics[width=\linewidth]{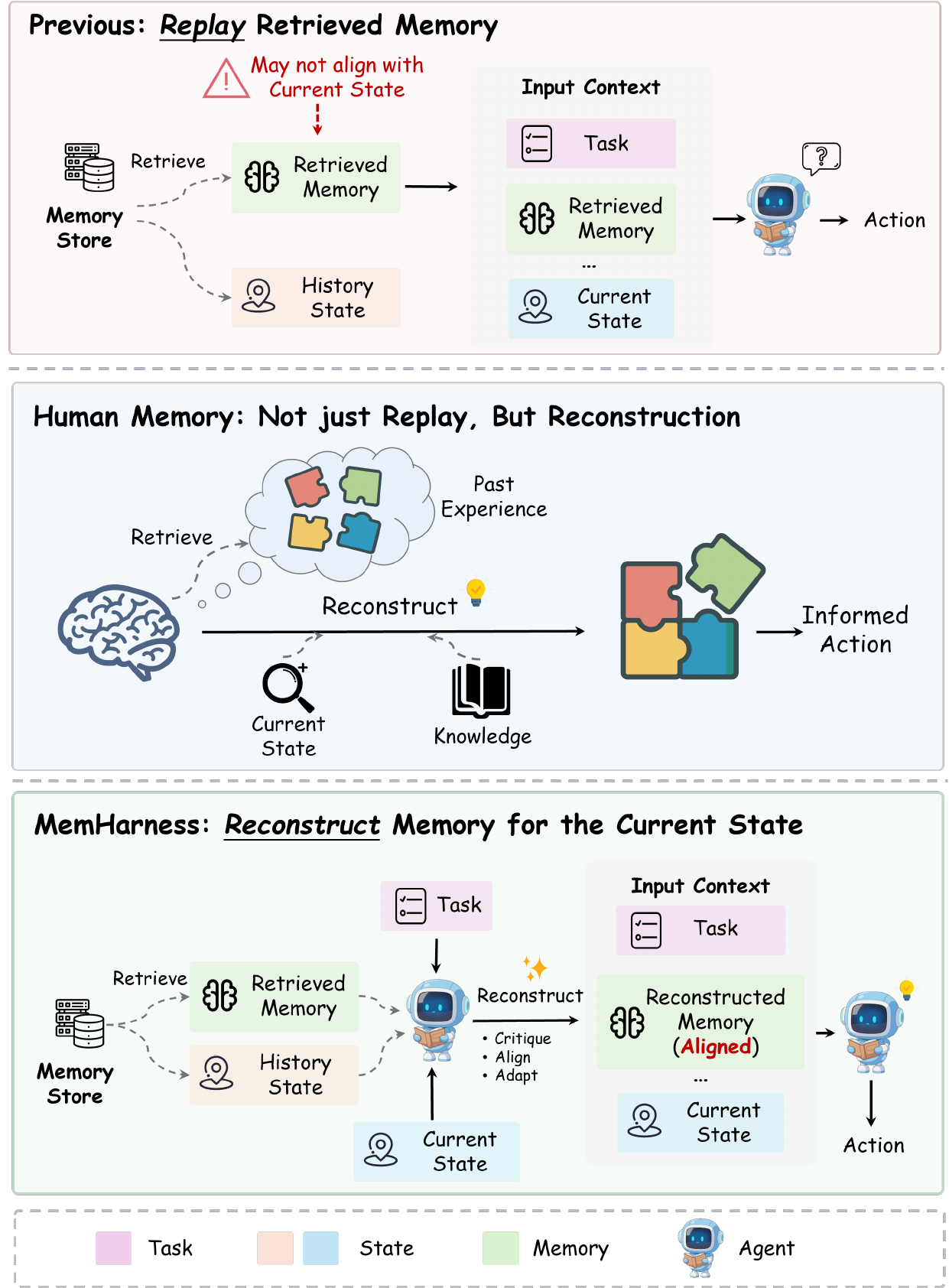}
    \caption{\textbf{Memory utilization paradigms.}
    \textbf{Top:} Prior methods directly replay retrieved memories, risking state misalignment.
    \textbf{Middle:} Human memory reconstructs past experience according to the current context.
    \textbf{Bottom:} Inspired by this process, MemHarness reconstructs retrieved memories into state-aligned guidance.}
    \label{fig:teaser}
\end{wrapfigure}

This middle ground echoes a fundamental insight from cognitive science: human remembering is reconstructive rather than reproductive~\cite{loftus1974reconstruction}. Recall does not recover an immutable record of the past; instead, retrieved experience is interpreted and reorganized using present cues and prior knowledge before it informs behavior. As illustrated in Figure~\ref{fig:teaser}, prior agents largely connect retrieval directly to action generation, whereas human memory introduces an intermediate process that evaluates and adapts recalled experience. This suggests a broader paradigm shift for memory-augmented agents---from \emph{retrieve and replay} to \emph{retrieve, evaluate, and reconstruct}.

Inspired by this mechanism, we decompose memory-guided decision-making into five stages that parallel the reconstructive process: \textbf{environment observation} captures the current situation; \textbf{experience retrieval} recalls a potentially relevant memory together with its historical context; \textbf{memory critique} assesses its applicability and identifies state mismatches; \textbf{contextual memory reconstruction} preserves transferable knowledge while revising or discarding incompatible content; and \textbf{action generation} uses the reconstructed guidance to make a context-aligned decision. 
Based on this formulation, we introduce \textbf{MemHarness}, designed to actively harness retrieved experiences by explicitly inserting critique and reconstruction between retrieval and action, turning memory from a static prompt fragment into context-sensitive guidance.

Building on this formulation, \textbf{MemHarness} bridges explicit and parametric memory utilization: it retains retrieved experiences as inspectable records, while parameterizing their state-conditioned critique and reconstruction within a single policy. At each decision step, conditioned on the current observation, agent first critiques the retrieved experience against its original context, then reconstructs it into state-specific guidance, and finally generates the executable action. Crucially, this reconstructive ability requires no additional human annotation: it emerges through end-to-end training with Group Relative Policy Optimization (GRPO). By optimizing for task success, the agent implicitly learns a discriminative reconstruction strategy, preserving transferable memories while rewriting misleading ones to close the gap between historical knowledge and the present state.

We evaluate MemHarness on two challenging agent decision-making benchmarks, ALFWorld and WebShop. It consistently outperforms both pure RL baselines and state-of-the-art static memory-augmented methods. Ablation studies further show that removing the reconstruction stage degrades performance to the level of naive memory replay, confirming that adaptive reconstruction, rather than retrieval alone, is the primary driver of the observed gains.

In summary, our main contributions are as follows:
\begin{itemize}[leftmargin=1.5em]
    \item \textbf{A reconstructive view of agent memory.} We identify the applicability gap underlying the prevalent \emph{verbatim replay} paradigm, and, drawing on the reconstructive nature of human memory, recast memory-guided decision-making as a five-stage process spanning observation, retrieval, critique, reconstruction, and action.

    \item \textbf{A reconstruction-centric framework.} We propose \textbf{MemHarness}, which actively harnesses retrieved experiences by inserting explicit memory reconstruction between retrieval and action, and learns this capability end-to-end via GRPO without external supervision, combining the traceability of explicit memory with the adaptivity of parameterized reconstruction.

    \item \textbf{Comprehensive empirical validation.} On interactive decision-making benchmarks, MemHarness delivers robust improvements over static replay, with ablations isolating reconstruction as the key contributing factor.
\end{itemize}

%% file: sections/related_work.tex
\section{Related Works}
\label{sec:related_works}

\subsection{LLM Agents for Interactive Decision-Making}
LLM agents navigate interactive tasks using prompting paradigms like ReAct~\cite{yao2023reactsynergizingreasoningacting, wei2022chain}, often augmented with planning or self-reflection~\cite{shinn2023reflexionlanguageagentsverbal, ma2026correctad}. However, these training-free methods are bounded by frozen model capabilities. To acquire task-specific skills from interaction, agents are fine-tuned via supervised learning~\cite{chen2024agent, chen2023fireact} or reinforcement learning (RL)~\cite{schulman2017proximal, ahmadian2024rloo, shao2024deepseekmath, feng2026group}. While RL optimizes decision-making directly, standard frameworks lack explicit mechanisms to accumulate and reuse cross-episode experiences, motivating memory-augmented agents.

\subsection{Memory-Augmented LLM Agents}
Memory-augmented agents address this by retrieving past interactions or distilled insights~\cite{zhao2024expelllmagentsexperiential, fang2026memp, chhikara2025mem0, liu2026simplemem}. Most rely on frozen policies, while alternative approaches internalize experience directly into model weights~\cite{zhang2025memgen}, sacrificing the traceability of explicit memory. Recent works integrate external memory with RL~\cite{zhang2026memrl, wu2025evolver}, yet they inject retrieved experiences \textit{verbatim}. Because interactive tasks exhibit high state variance, replay often introduces misaligned or misleading guidance. 

MemHarness bridges these two paradigms. It retains explicit memory banks for traceability but parameterizes their state-conditioned reconstruction. Rather than applying retrieved memory blindly, MemHarness is optimized end-to-end to critique and rewrite past experiences into context-specific guidance before acting.

%% file: sections/methodology.tex
\section{Method}
\label{method}

\begin{figure*}[t]
    \centering
    \includegraphics[width=\linewidth]{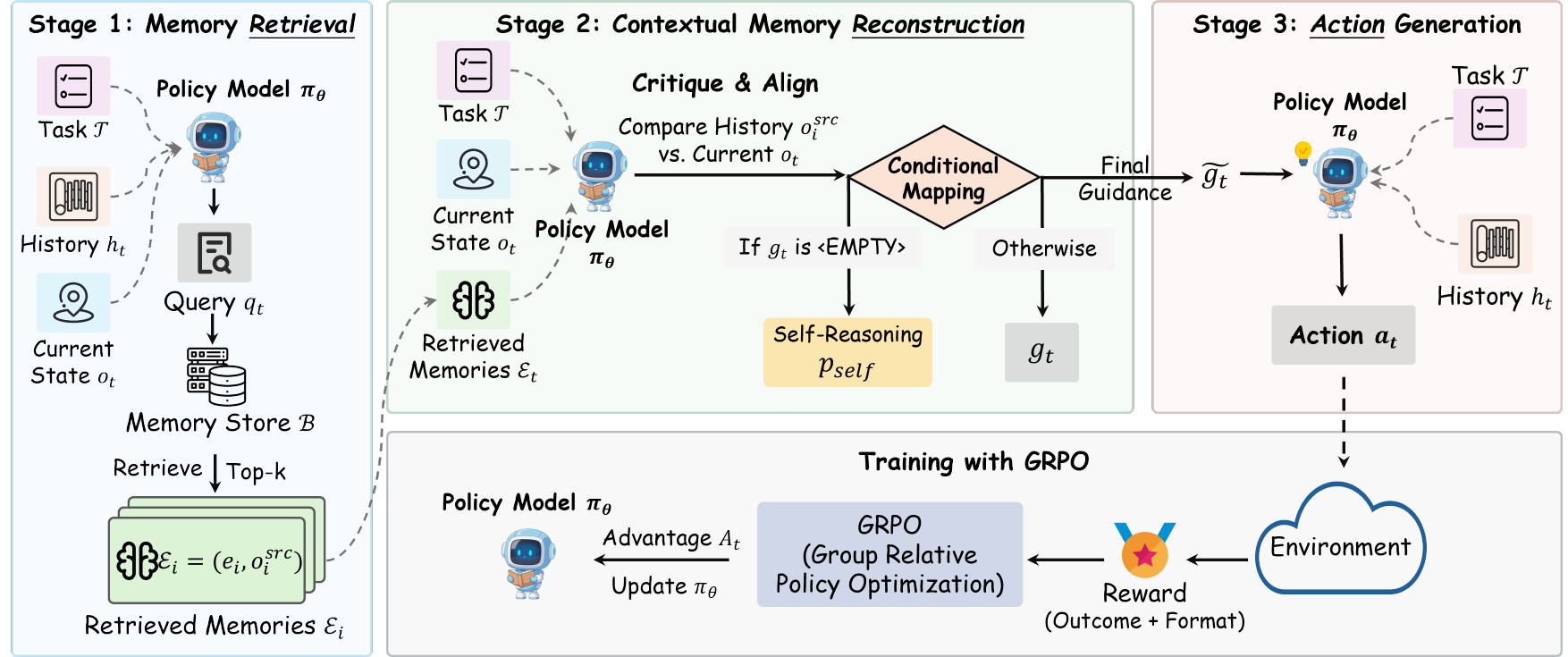}
    \caption{Overview of the MemHarness framework. The execution pipeline consists of three stages: (1) \textbf{Memory Retrieval}, where the policy generates a query to fetch relevant past experiences; (2) \textbf{Contextual Memory Reconstruction}, where the policy compares the memory's source state with the current state to reconstruct adapted guidance (or falls back to self-reasoning if the memory is deemed unhelpful); and (3) \textbf{Action Generation} guided by the reconstructed memory.}
    \label{fig:framework}
\end{figure*}

\subsection{Problem Formulation}
\label{sec:problem}

\paragraph{Sequential Decision-Making.}
We model interactive decision-making as a POMDP
$\mathcal{M}=(\mathcal{S},\mathcal{A},\mathcal{O},P,R)$, where
$\mathcal{S}$, $\mathcal{A}$, and $\mathcal{O}$ are the state, action, and
observation spaces, respectively, and $P$ and $R$ denote the transition and
reward functions. Given a task description $\mathcal{T}$, at step $t$ the
agent observes $o_t$ and selects $a_t$ using a policy $\pi_\theta$. To account
for partial observability, the policy conditions on a recent history window
$h_t=(o_{t-w+1},a_{t-w+1},\ldots,o_{t-1},a_{t-1},o_t)$. The objective is to
maximize the expected episodic return
$\mathbb{E}_{\pi_\theta}[\sum_{t=1}^{T}R(s_t,a_t)]$.

\paragraph{Experiential Memory.}
The agent maintains a memory bank $\mathcal{B}=\{m_i\}_{i=1}^{N}$ by
periodically distilling collected trajectories into natural-language
experiences. Each entry $m_i=(e_i,o_i^{src})$ contains an abstract strategy
$e_i$ and its source observation $o_i^{src}$. Following~\cite{wu2025evolver},
we perform semantic deduplication, track empirical utility and periodically
prune low-utility entries. Given a context-derived query $q_t$, a retriever
$\mathcal{R}$ returns the top-$k$ entries:
\begin{equation}
    \label{eq:retrieval}
    \mathcal{E}_t=\mathcal{R}(q_t,\mathcal{B}),
    \qquad |\mathcal{E}_t|=k.
\end{equation}
The source observations enable comparison between historical and current
states during reconstruction.

\paragraph{From Replay to Reconstruction.}
Verbatim replay directly conditions action generation on retrieved experiences:
$a_t\sim\pi_\theta(\cdot\mid\mathcal{T},h_t,\mathcal{E}_t)$. MemHarness instead
first reconstructs them into state-specific guidance and then acts:
\begin{equation}
\begin{aligned}
    g_t &\sim \pi_\theta(\cdot\mid\mathcal{T},h_t,\mathcal{E}_t),\\
    \tilde{g}_t &= f(g_t),\\
    a_t &\sim \pi_\theta(\cdot\mid\mathcal{T},h_t,\tilde{g}_t),
\end{aligned}
\end{equation}
where $f$ maps the raw reconstruction $g_t$ to final guidance
$\tilde{g}_t$, including rejection of mismatched memories. Both stages share
the unified policy $\pi_\theta$.

\subsection{Overall Framework}
\label{sec:framework}

We introduce \textbf{MemHarness}, an RL-driven framework that reconstructs retrieved experiences for the current state rather than replaying them verbatim. As shown in Figure~\ref{fig:framework}, its inference pipeline comprises three stages: retrieval, contextual memory reconstruction, and action generation. These three stages instantiate the five-stage formulation in Section~\ref{sec:intro}: the history window $h_t$ serves as the environment observation, and memory critique is folded into the reconstruction stage.

\begin{enumerate}[left=1em]
    \item \textbf{Retrieval.} At step $t$, the agent decides whether to query the experiential memory bank $\mathcal{B}$ based on $h_t$. If triggered, retrieval returns relevant experiences $\mathcal{E}_t$, which are reconstructed rather than directly added to the action context.

    \item \textbf{Reconstruction.} The policy $\pi_\theta$ critiques $\mathcal{E}_t$ against $h_t$ and generates state-specific guidance $g_t$, retaining applicable knowledge while discarding or revising misaligned content.

    \item \textbf{Action Generation.} The same policy $\pi_\theta$ generates an executable action $a_t$ conditioned on $h_t$ and $g_t$.
\end{enumerate}

MemHarness thus separates reconstruction from action generation while implementing both with a unified policy. Because supervised reconstruction traces are unavailable, we train the two-step process end-to-end with GRPO using only task rewards. The following sections detail the reconstruction and optimization procedures.

\subsection{Contextual Memory Reconstruction}
\label{sec:reconstruction}

MemHarness transforms retrieved experiences into state-specific guidance through an explicit reconstruction. Because experiences abstracted from historical trajectories may be outdated or incompatible with the current state, the policy critiques their applicability before they influence action generation.

\paragraph{Critique and Reconstruction.}
The reconstruction context concatenates the task description $\mathcal{T}$, current history $h_t$, and retrieved memory tuples $(e_{t,i}, o_{t,i}^{src}) \in \mathcal{E}_t$:
\begin{equation}
    x_{\mathrm{recon}}
    = \mathcal{T} \oplus h_t
    \oplus \bigcup_{i=1}^{k}(e_{t,i}, o_{t,i}^{src}),
\end{equation}
where $e_{t,i}$ is a retrieved experience, $o_{t,i}^{src}$ is its source observation, and $\oplus$ denotes text concatenation. The policy then generates:
\begin{equation}
    g_t \sim \pi_\theta(\cdot \mid x_{\mathrm{recon}}).
\end{equation}
By comparing each source observation with the current context represented by $h_t$, the policy identifies state shifts and retains, revises or rejects the retrieved experience to produce context-specific guidance.

If no retrieved experience is applicable, the policy outputs
\texttt{<EMPTY>}. The guidance used for action generation is
\begin{equation}
    \tilde{g}_t =
    \begin{cases}
        p_{\mathrm{self}}, & \text{if } g_t=\texttt{<EMPTY>},\\
        g_t, & \text{otherwise},
    \end{cases}
\end{equation}
where $p_{\mathrm{self}}$ instructs the agent to rely on its own reasoning without memory guidance.

\paragraph{Action Generation.}
The same policy generates the executable action conditioned on the reconstructed guidance:
\begin{equation}
    a_t \sim \pi_\theta(\cdot \mid \mathcal{T}, h_t, \tilde{g}_t).
\end{equation}
Thus, action generation uses either adapted memory or independent reasoning, rather than unfiltered historical records. Reconstruction and action generation share parameters $\theta$ and are optimized jointly. Since $g_t$ has no ground-truth annotation, we treat it as a latent reasoning process and train the pipeline end-to-end using task-level rewards

\subsection{Training with GRPO}
\label{sec:training}
Since the guidance $g_t$ is a latent variable without ground-truth supervision, we optimize the retrieve--reconstruct--act pipeline end-to-end using reinforcement learning. The reward combines task outcomes with format-based signals that enforce valid reconstruction and action outputs. We adopt GRPO to estimate advantages from groups of sampled trajectories without training a separate value network.

\paragraph{Reward Design.}
The reward $R(\tau_i)$ for a trajectory primarily depends on the sparse task outcome (0 for failure, 10 for success), augmented by a minor bonus to enforce structural constraints: 
$R(\tau_i) = R_{\text{outcome}} + 0.1 \times R_{\text{format}}$. 
The format score $R_{\text{format}} \in [0, 1]$ equally weights three criteria: (1) exactly one valid \thinktag{think} and \actiontag{action} block per step, (2) appropriate memory retrieval frequency with \retrievetag{retrieve\_memory} block (1 to 5 times per episode), and (3) strict English output. This formatting bonus explicitly encourages an agentic protocol (i.e., reason first, retrieve memory when necessary, single action per step) without overshadowing the primary task objective.

\paragraph{Policy Update.}
For each task, we sample a group of $G$ trajectories
$\{\tau_i\}_{i=1}^{G}$. GRPO computes the group-normalized advantage:
\begin{equation}
    A_i =
    \frac{R(\tau_i)-\mathrm{mean}(\{R(\tau_k)\}_{k=1}^{G})}
         {\mathrm{std}(\{R(\tau_k)\}_{k=1}^{G})}.
\end{equation}
The trajectory-level advantage $A_i$ is assigned to all tokens in $\tau_i$,
providing shared credit to thinking process, reconstruction and action generation. For token
$y_{i,j}$, the importance ratio is
\begin{equation}
    r_{i,j}(\theta)=
    \frac{\pi_\theta(y_{i,j}\mid y_{i,<j})}
         {\pi_{\theta_{\mathrm{old}}}(y_{i,j}\mid y_{i,<j})}.
\end{equation}
The clipped surrogate objective is
\begin{equation}
    \mathcal{L}^{\mathrm{CLIP}}_{i,j}(\theta)
    =\min\!\left(
        r_{i,j}(\theta)A_i,\,
        \mathrm{clip}(r_{i,j}(\theta),1-\epsilon,1+\epsilon)A_i
    \right),
\end{equation}
and the overall objective is
\begin{equation}
    \mathcal{J}(\theta)=\mathbb{E}\!\left[
    \frac{1}{\sum_i|\tau_i|}
    \sum_{i=1}^{G}\sum_{j=1}^{|\tau_i|}
    \left(
        \mathcal{L}^{\mathrm{CLIP}}_{i,j}(\theta)
        -\beta\,\mathbb{D}_{\mathrm{KL}}
        [\pi_\theta\|\pi_{\mathrm{ref}}]
    \right)\right],
\end{equation}
where $\epsilon$ controls clipping and $\beta$ weights the KL regularization.
This objective jointly optimizes memory reconstruction and action generation
using the same trajectory-level reward.

%% file: sections/experiments.tex
\section{Experiments}
\label{sec:experiments}
In this section, we empirically evaluate our proposed MemHarness framework on long-horizon interactive environments. Our evaluation is designed to answer the following core research questions (RQs):
\begin{itemize}[leftmargin=1.5em]
    \item \textbf{RQ1:} Does the MemHarness framework outperform existing agentic and memory-augmented baselines on complex interactive tasks? (Section~\ref{sec:main_results})
    \item \textbf{RQ2:} How do individual components (especially the contextual memory reconstruction module) contribute to the overall performance? (Section~\ref{sec:ablations})
    \item \textbf{RQ3:} Is MemHarness robust against environmental distribution shifts in out-of-distribution (OOD) scenarios? (Section~\ref{sec:ood})
    \item \textbf{RQ4:} Does the reconstruction mechanism genuinely rely on state comparison to make decisions? (Section~\ref{sec:mechanism})
    \item \textbf{RQ5:} How do adaptive memory utilization behaviors evolve during reinforcement learning, and how do they correlate with task success? (Section~\ref{sec:train_dynamic})
\end{itemize}

\subsection{Experimental Setup}
\label{sec:exp_setup}

\paragraph{Benchmarks and Baselines.}
We evaluate MemHarness on two complex, long-horizon interactive benchmarks: ALFWorld~\cite{shridhar2020alfworld} (embodied household tasks) and WebShop~\cite{yao2022web} (goal-directed online shopping). Performance is measured by task success rate, together with the average task score on WebShop. We compare MemHarness against closed-source LLMs, including GPT-4o~\cite{hurst2024gpt} and Gemini-2.5-Pro~\cite{comanici2025gemini}, prompt-based or memory-based agents (ReAct, Reflexion, Mem0, ExpeL, MemP, and SimpleMem)~\cite{yao2023reactsynergizingreasoningacting, shinn2023reflexionlanguageagentsverbal, chhikara2025mem0, zhao2024expelllmagentsexperiential, fang2026memp, liu2026simplemem}, and RL-trained agents (RLOO, GRPO, MemRL, EvolveR, Mem0+GRPO, and SimpleMem+GRPO)~\cite{ahmadian2024rloo, guo2025deepseek, zhang2026memrl, wu2025evolver, liu2026simplemem}. We also evaluate ablated variants with verbatim memory replay or the reconstruction harness removed at inference time to isolate the contribution of our design.

\paragraph{Implementation Details.}
We adopt Qwen2.5-7B-Instruct~\cite{qwen2025qwen25technicalreport} as the policy backbone. Our implementation builds on the \texttt{verl-agent} framework~\cite{feng2026group}. Following Deepseek-R1 and EvolveR~\cite{guo2025deepseek, wu2025evolver}, we perform a brief cold-start stage before RL training. For each benchmark, the cold-start dataset contains 200 multi-turn interaction trajectories with chain-of-thought reasoning and active retrieval of memories, as well as 200 question--answer examples that summarize trajectories into memories. Whenever memory is retrieved, the agent is required to reconstruct it before taking an action. This stage primarily aligns the model with the required interaction and memory-abstraction formats. The unified policy is subsequently optimized end-to-end via GRPO with sparse task rewards ($10$ for success and $0$ for failure) and a minor formatting bonus. For fair comparison, all trainable baselines use the same backbone, observation history window ($w=3$), interaction budget, and optimization configuration. We will show more details in the Appendix.

\paragraph{Memory Configuration.}
The experiential memory bank is initialized as empty and progressively populated during RL training with memory principles summarized by the policy from its own interaction trajectories. We use BGE-M3~\cite{chen2024bge} as the embedding model and retrieve the top-$3$ most relevant principles.

\input{table/main_result}

\subsection{Main Results: Overall Effectiveness} 
\label{sec:main_results}

Table~\ref{tab:main_results} summarizes the performance on AlfWorld and WebShop. Overall, MemHarness achieves the best results, reaching success rates of 85.2\% on AlfWorld and 75.6\% on WebShop. 

\textbf{Comparison with Foundation Models.} 
MemHarness demonstrates massive improvements over its base model, Qwen2.5-7B-Instruct, and substantially outperforms strong closed-source models like Gemini-2.5-Pro (+23.1\% on AlfWorld and +39.7\% on WebShop) despite its 7B parameter scale.

\textbf{Advantage over Existing Paradigms.} 
Prompt-based and test-time memory methods (e.g., Reflexion, ExpeL) yield limited success without policy optimization. While pure RL (GRPO) provides a strong baseline, naively combining it with external memory (e.g., Mem0+GRPO, SimpleMem+GRPO) severely degrades performance. MemHarness effectively overcomes this issue via state-conditioned memory reconstruction, safely leveraging historical experience to outperform standard GRPO by absolute margins of +8.8\% and +9.5\% on AlfWorld and WebShop, respectively.

\subsection{Ablation Studies: Component Analysis}  
\label{sec:ablations}

To evaluate the contribution of each component (RQ2), we conduct ablation studies on ALFWorld and WebShop, with results summarized in Table~\ref{tab:ablation}. We identify three key findings regarding memory utilization and policy optimization:

\paragraph{Necessity of RL.} 
The cold-start model, trained solely for format alignment, performs poorly (7.6\% on ALFWorld), confirming that RL is necessary for acquiring actual interactive decision-making capabilities.

\paragraph{Raw Memory Introduces Noise.} 
Injecting unrefined memory (``RL + Raw Memory'' or ``w/o reconstruction'') hurts ALFWorld performance. This proves verbatim replay introduces state-mismatch noise, which our reconstruction mechanism effectively mitigates.

\paragraph{Intrinsic Policy Improvement.} 
Disabling test-time memory (``w/o memory'') still substantially outperforms the `RL Only` baseline (83.0\% vs. 76.4\% on ALFWorld). Thus, the reconstruction objective during training intrinsically enhances the agent's base reasoning capabilities, serving as high-quality latent guidance.

\paragraph{Necessity of Policy-Internal Reconstruction.} 
Replacing MemHarness's internal reconstruction module with a generic instruction-tuned LLM (\textit{Qwen2.5-7B-Instruct}) while keeping the actor unchanged degrades performance significantly (e.g., 85.2\% to 77.7\% on ALFWorld). This demonstrates that zero-shot text rewriting is insufficient; end-to-end RL is crucial for learning task-grounded adaptation that aligns with environment dynamics.

\input{table/ablation_result}

\subsection{Out-of-Distribution Generalization}
\label{sec:ood}

To address RQ3, we evaluate robustness against environmental shifts on ALFWorld Out-of-Distribution (OOD) settings, where room layouts and object placements are unseen during training. In such scenarios, directly applying historical memories risks severe state mismatch.

As shown in Table~\ref{tab:ood_alfworld}, MemHarness achieves the highest average success rate (85.9\%), substantially outperforming the ``RL + Raw Memory'' baseline (76.3\%). Notably, while the intrinsic policy (``w/o memory'') remains competitive (83.0\%), injecting unadapted memories (``w/o reconstruction'') degrades performance to 82.4\%. This confirms that verbatim memory replay in unseen environments introduces noise. MemHarness overcomes this by dynamically filtering and rewriting mismatched guidance, safely leveraging past experiences to achieve superior OOD generalization.

\input{table/alfworld_ood}

\begin{figure*}[t]
    \centering
    \begin{minipage}{0.49\linewidth}
        \centering
        \includegraphics[width=\linewidth]{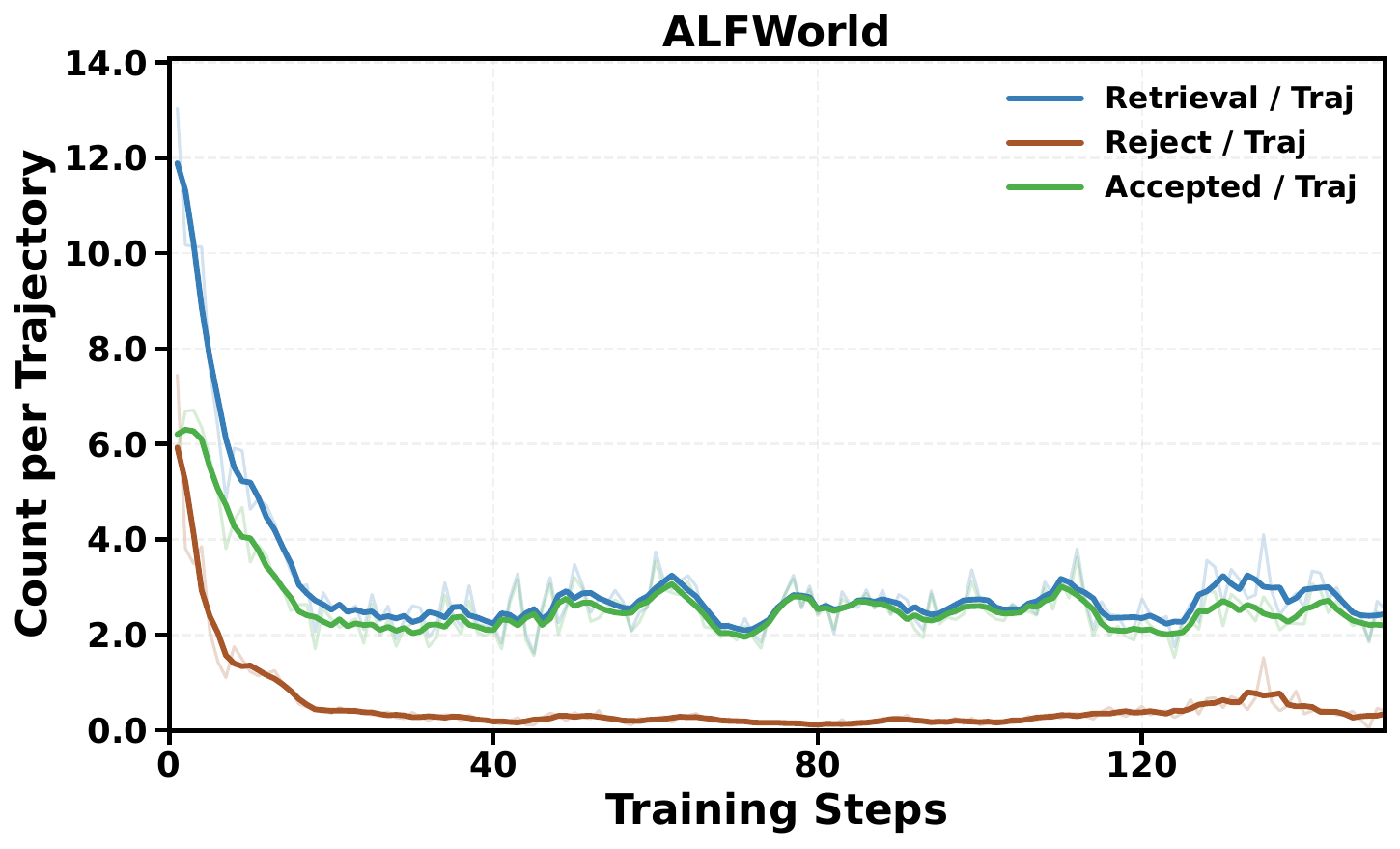}
        \caption*{ (a) ALFWorld }
    \end{minipage}
    \hfill
    \begin{minipage}{0.49\linewidth}
        \centering
        \includegraphics[width=\linewidth]{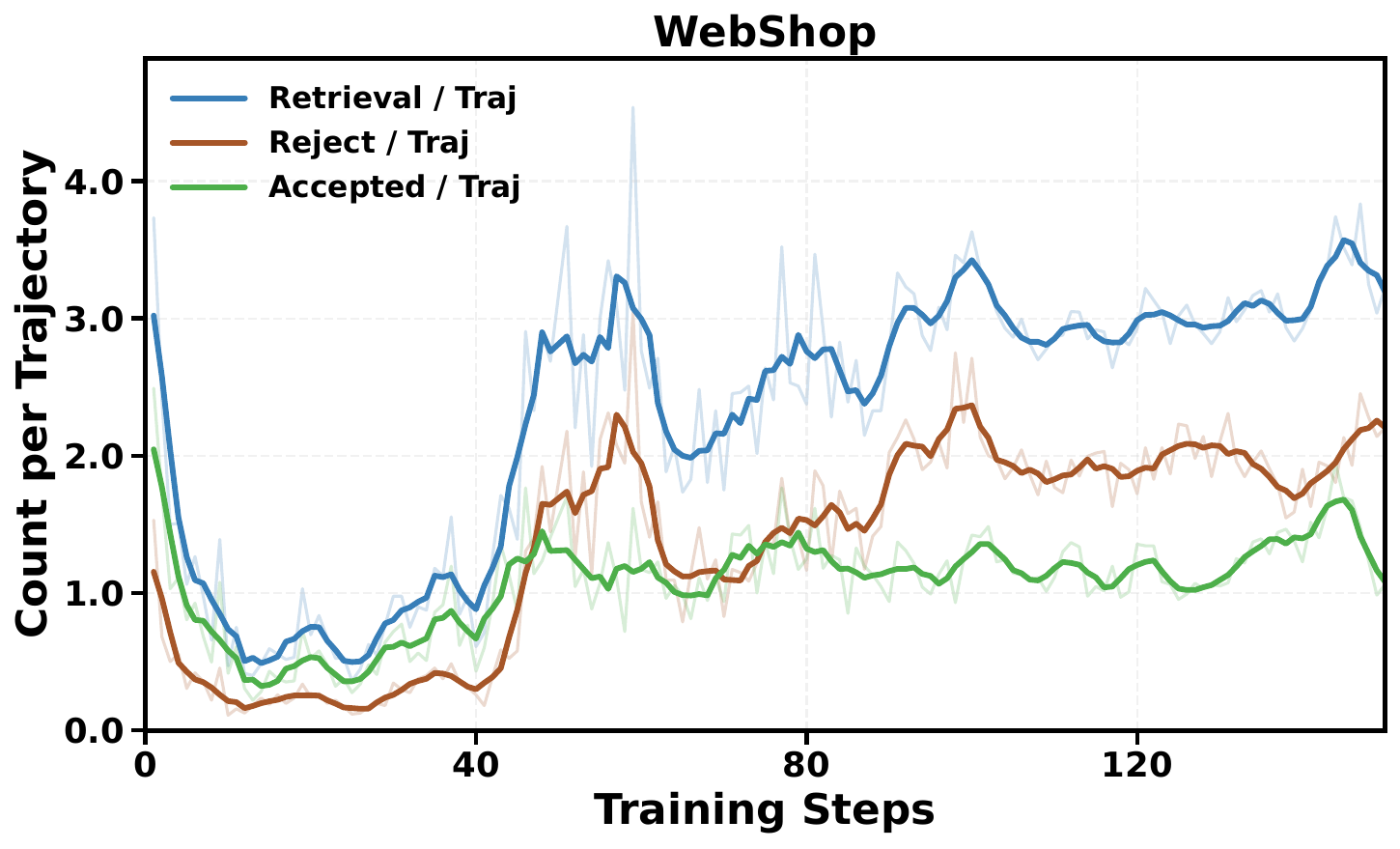}
        \caption*{ (b) WebShop }
    \end{minipage}
    \caption{\textbf{Evolution of memory usage behaviors during RL training.} The curves show the average number of memory retrievals, accepted reconstructions, and rejected reconstructions per trajectory.}
    \label{fig:behavior_count}
\end{figure*}

\begin{figure*}[!ht]
    \centering
    \begin{minipage}{0.49\linewidth}
        \centering
        \includegraphics[width=\linewidth]{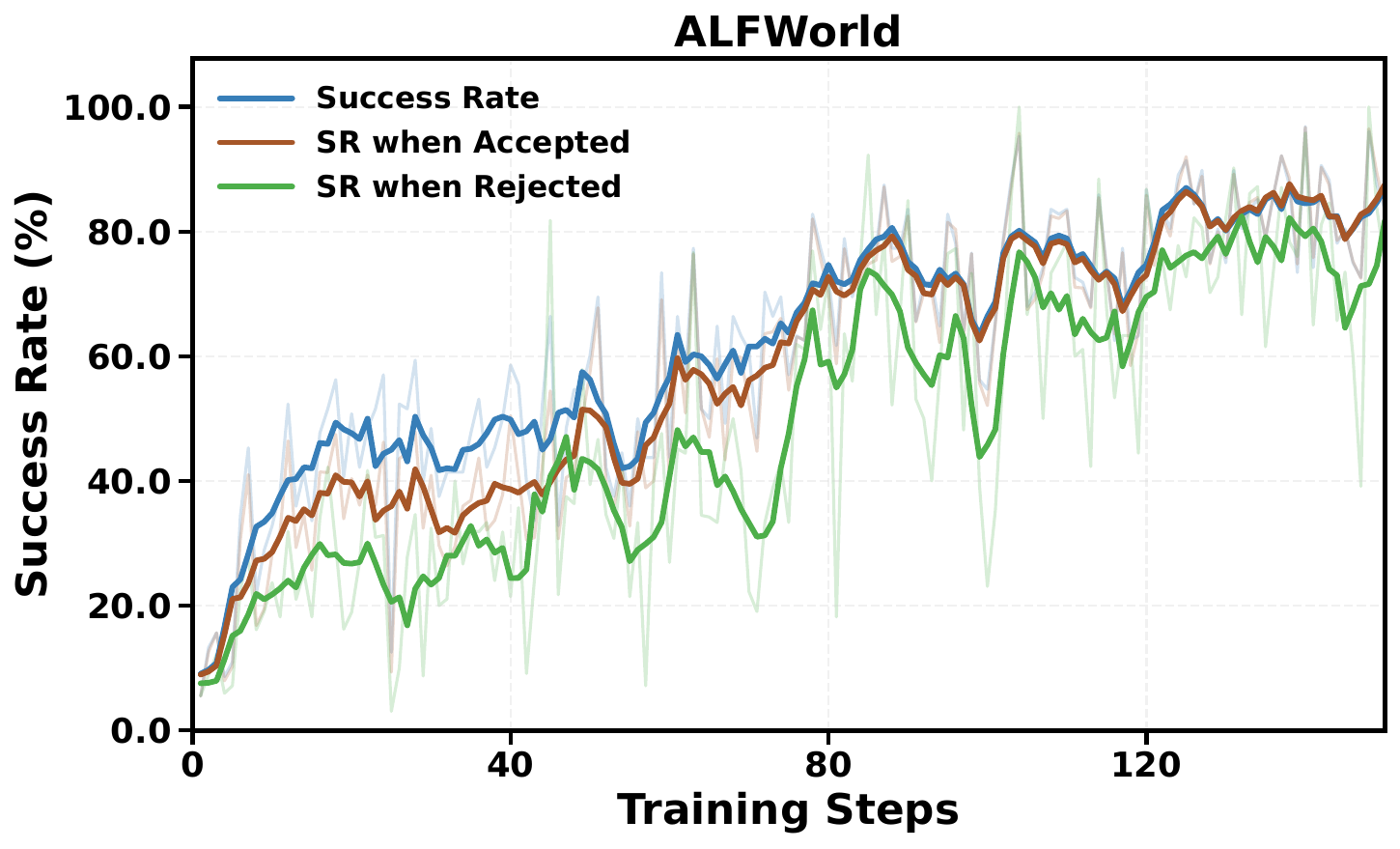}
        \caption*{ (a) ALFWorld }
    \end{minipage}
    \hfill
    \begin{minipage}{0.49\linewidth}
        \centering
        \includegraphics[width=\linewidth]{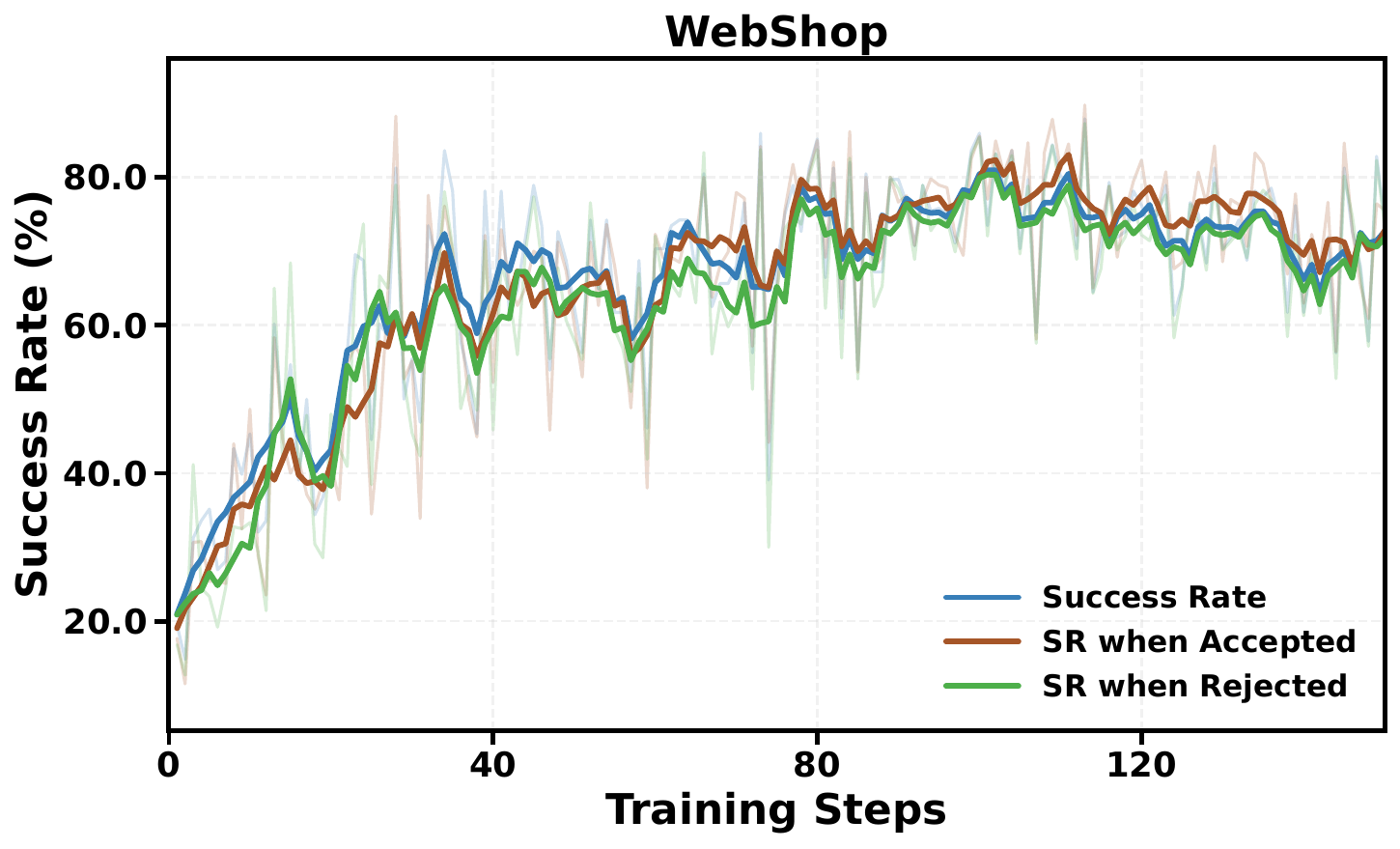}
        \caption*{ (b) WebShop }
    \end{minipage}
    \caption{\textbf{Training dynamics of success rates conditioned on memory decisions.} ``SR when Accepted/Rejected'' denotes the success rate of trajectories containing at least one accepted or rejected memory reconstruction.}
    \label{fig:behavior_sr}
\end{figure*}

\subsection{Mechanism Deep Dive: How is Memory Reconstructed?}
\label{sec:mechanism}

To answer RQ4, we investigate whether reconstruction genuinely relies on state comparison, examining its behavior from macro-level filtering to micro-level adaptation.

\paragraph{Macro-level Filtering.}
We ablate the historical source state ($o^{src}$) provided during reconstruction (Table~\ref{tab:reconstruction_analysis}). Removing $o^{src}$ entirely leaves the rejection rate unchanged but degrades task success (e.g., $85.2\% \to 80.0\%$ on ALFWorld), as the agent accepts misaligned guidance. Crucially, replacing $o^{src}$ with a mismatched state from a \textit{random} memory markedly increases rejection rates ($8.7\% \to 13.3\%$ on ALFWorld, $56.0\% \to 63.3\%$ on WebShop). This confirms the policy actively compares historical and current states to filter incompatible memories, rather than reacting to missing fields, thereby maintaining robust downstream performance.

\paragraph{Micro-level Counterfactual Adaptation.}
We construct 1,000 offline probes by asking a strong LLM to minimally edit each current state so that its retrieved memory becomes inapplicable, and evaluate only the reconstruction output rather than task execution (Appendix~\ref{sec:app_counterfactual}). Table~\ref{tab:reconstruction_analysis} reports the resulting within-environment shifts. In ALFWorld, counterfactual edits reduce unchanged outputs and increase adaptation and rejection. In WebShop, they further increase rejection. The different response patterns are consistent with their observation formats: ALFWorld provides concise state descriptions, whereas WebShop exposes substantially longer webpage content. Despite this difference, both show that reconstruction responds to fine-grained state changes.

\input{table/mechanism_res}

\subsection{Training Dynamics and Behavioral Evolution (RQ5)}
\label{sec:train_dynamic}

To address RQ5, we track internal memory decisions and task success rates during RL training. MemHarness learns task-specific memory strategies rather than fixed rules (Figure~\ref{fig:behavior_count}). ALFWorld converges to a sparse policy (retrievals dropping to 2--3 per trajectory), whereas WebShop maintains frequent retrieval with persistent active filtering. These learned decisions align directly with task outcomes (Figure~\ref{fig:behavior_sr}). Across both environments, trajectories containing \textit{accepted} reconstructions consistently achieve higher success rates than those with \textit{rejected} ones, confirming the agent effectively learns to judge memory applicability.



%% file: table/main_result.tex
\begin{table*}[t]
\centering
\setlength{\tabcolsep}{8.2pt}
\begin{tabular}{l*{7}{c}|*{2}{c}} 
\toprule
\multirow{2}{*}{\textbf{Method}} & \multicolumn{7}{c|}{\textbf{AlfWorld}} & \multicolumn{2}{c}{\textbf{WebShop}} \\
\cmidrule(lr){2-8} \cmidrule(lr){9-10}
 & Pick & Look & Clean & Heat & Cool & Pick2 & Avg. SR & Score & SR \\
\midrule
\multicolumn{10}{l}{\textit{Closed-source LLMs}} \\
GPT-4o          & 75.3 & 60.8 & 31.2 & 56.7 & 21.6 & 49.8 & 49.2 & 31.8 & 23.7 \\
Gemini-2.5-Pro  & \textbf{92.8} & 63.3 & 62.1 & 69.0 & 26.6 & 58.7 & 62.1 & 42.5 & 35.9 \\
\midrule
\multicolumn{10}{l}{\textit{Qwen2.5-7B-Instruct}} \\
Qwen2.5         & 33.4 & 21.6 & 19.3 & 6.9  & 2.8  & 3.2  & 14.5 & 26.4 & 7.8 \\
\midrule
\multicolumn{10}{l}{\textit{Prompt-based Agentic or Memory-based Methods}} \\
ReAct     & 48.5 & 35.4 & 34.3 & 13.2 & 18.2 & 17.6 & 27.9 & 46.2 & 19.5 \\
Reflexion & 62.0 & 41.6 & 44.9 & 30.9 & 36.3 & 23.8 & 39.9 & 58.1 & 28.8 \\
Mem0            & 54.0 & 55.0 & 26.9 & 36.4 & 20.8 & 7.7  & 33.5 & 23.9 & 2.0 \\
ExpeL           & 21.0 & 67.0 & 55.0 & 52.0 & 71.0 & 6.0  & 45.3 & 30.9 & 11.2 \\
MemP            & 54.3 & 38.5 & 48.1 & 56.2 & 32.0 & 16.7 & 41.0 & 25.3 & 6.4 \\
SimpleMem       & 64.5 & 33.3 & 20.0 & 12.5 & 33.3 & 3.8  & 27.9 & 33.2 & 8.6 \\
\midrule
\multicolumn{10}{l}{\textit{RL-based Methods}} \\
RLOO      & 87.6 & 78.2 & 87.3 & 81.3 & 71.9 & 48.9 & 75.9 & 80.3 & 65.7 \\
GRPO      & 90.8 & 66.1 & 89.3 & 74.7 & \textbf{72.5} & 64.7 & 76.4 & 79.3 & 66.1 \\
\midrule
\multicolumn{10}{l}{\textit{Memory-Augmented RL-based Methods}} \\
MemRL           & 62.8 & 38.5 & 22.2 & 12.5 & 8.0  & 0.0  & 24.0 & 29.5 & 9.2 \\
Mem0+GRPO       & 78.1 & 54.8 & 56.1 & 31.0 & 65.0 & 26.9 & 52.0 & 58.1 & 37.5 \\
SimpleMem+GRPO  & 89.5 & 36.3 & 60.0 & 50.0 & 64.9 & 26.3 & 54.5 & 67.8 & 46.9 \\
EvolveR (reproduced)  & 84.8 & 61.5 & 95.7 & 62.5 & 61.9 & 54.2 & 70.1 & 84.1 & 72.6 \\
\rowcolor{gray!15}
\textbf{MemHarness} & 87.0 & \textbf{78.6} & \textbf{97.0} & \textbf{87.5} & 71.4 & \textbf{90.0} & \textbf{85.2} & \textbf{87.4} & \textbf{75.6} \\
\bottomrule
\end{tabular}
\caption{Main results on AlfWorld and WebShop. For AlfWorld, we report the success rate (\%) on each of the six task categories, along with the macro-averaged success rate (Avg. SR) across all categories. Best results are highlighted in \textbf{bold}.}
\label{tab:main_results}
\end{table*}

%% file: table/ablation_result.tex
\begin{table*}[t]
\centering
\small
\setlength{\tabcolsep}{3.1pt}
\begin{tabular}{l cc cccccc c c}
\toprule
\multirow{2}{*}{\textbf{Method}} & \multicolumn{2}{c}{\textbf{Test-time}} & \multicolumn{7}{c}{\textbf{ALFWorld}} & \textbf{WebShop} \\
\cmidrule(lr){2-3} \cmidrule(lr){4-10} \cmidrule(lr){11-11}
 & Memory & Reconstruction & Pick & Look & Clean & Heat & Cool & Pick2 & Avg. SR & Avg. SR \\
\midrule
Base Model               & \xmark & \xmark & 33.4 & 21.6 & 19.3 & 6.9 & 2.8 & 3.2 & 14.5 &  7.8 \\
Cold Start Model         & \xmark & \xmark & 11.8 & 21.1 & 0 & 7.1 & 0 & 5.3 & 7.6  &  17.6 \\
RL Only (GRPO)           & \xmark & \xmark & 90.8 & 66.1 & 89.3 & 74.7 & 72.5 & 64.7 & 76.4 &  66.1 \\
RL + Raw Memory          & \cmark & \xmark & 84.8 & 61.5 & 95.7 & 62.5 & 61.9 & 54.2 & 70.1 &  72.6 \\
\midrule
\rowcolor{gray!15}
\textbf{MemHarness (Ours)}  & \cmark & \cmark & 87.0 & 78.6 & 97.0 & 87.5 & 71.4 & 90 & \textbf{85.2} & \textbf{75.6} \\
\quad -- generic LLM reconstrucion & \cmark & \cmark & 97.1 & 52.6 & 79.2 & 78.6 & 80 & 78.9 & 77.7  & 71.8 \\
\quad -- w/o reconstruction & \cmark & \xmark & 94.1 & 52.6 & 79.2 & 85.7 & 76.7 & 89.5 & 79.6  & 74.6 \\
\quad -- w/o memory         & \xmark & \xmark & 100 & 68.4 & 91.7 & 78.6 & 70 & 89.5 & 83.0  & 73.6 \\
\bottomrule
\end{tabular}
\caption{Ablation studies on the components of MemHarness. Memory and Reconstruction indicate whether memory retrieval and reconstruction modules are active during inference. Best results are in \textbf{bold}.}
\label{tab:ablation}
\end{table*}

%% file: table/alfworld_ood.tex
\begin{table}[t]
\centering
\begin{tabularx}{\linewidth}{@{}l*{7}{C}@{}}
\toprule
\multirow{2}{*}{\textbf{Method}} & \multicolumn{7}{c}{\textbf{ALFWorld OOD Success Rate (\%)}} \\
\cmidrule(l){2-8}
 & Pick & Look & Clean & Heat & Cool & Pick2 & \textbf{Avg. SR} \\
\midrule
\rowcolor{gray!15}
\textbf{MemHarness} & 97.1 & \textbf{73.7} & 87.5 & \textbf{92.9} & 80.0 & 84.2 & \textbf{85.9} \\
\quad -- w/o reconstruction & 94.1 & 52.6 & 87.5 & \textbf{92.9} & 83.3 & 84.2 & 82.4 \\
\quad -- w/o memory & \textbf{100.0} & 68.4 & \textbf{91.7} & 78.6 & 70.0 & \textbf{89.5} & 83.0 \\
\midrule
RL + Raw Memory & 91.2 & 68.4 & 75.0 & 78.6 & \textbf{86.7} & 57.9 & 76.3 \\
\bottomrule
\end{tabularx}
\vspace{1em}
\caption{Performance on the ALFWorld Out-of-Distribution (OOD) evaluation. The environment layouts and object configurations are unseen during training.}
\label{tab:ood_alfworld}
\end{table}


%% file: table/mechanism_res.tex
\begin{table*}[t]
\centering
\small

\begin{minipage}[t]{0.49\textwidth}
\centering
\textbf{(a) Source-state ablation}\par\vspace{3pt}
{\renewcommand{\arraystretch}{1.35} 
\setlength{\tabcolsep}{3.5pt}
\begin{tabular}{@{}lcccc@{}}
\toprule
\multirow{2}{*}{Source State}
& \multicolumn{2}{c}{ALFWorld}
& \multicolumn{2}{c}{WebShop} \\
\cmidrule(lr){2-3}\cmidrule(l){4-5}
& RR (\%) & SR (\%) & RR (\%) & SR (\%) \\
\midrule
Correct source ($o^{src}$) & 8.7  & 85.2 & 56.0 & 75.6 \\
No source (w/o $o^{src}$)  & 7.8  & 80.0 & 55.9 & 73.0 \\
Random source              & 13.3 & 84.3 & 63.3 & 73.6 \\
\bottomrule
\end{tabular}
}
\end{minipage}
\hfill
\begin{minipage}[t]{0.49\textwidth}
\centering
\textbf{(b) Counterfactual behavior}\par\vspace{3pt}

\setlength{\tabcolsep}{3.5pt}
\begin{tabular}{@{}llccc@{}}
\toprule
\multirow{2}{*}{Task}
& \multirow{2}{*}{State}
& \multicolumn{3}{c}{Output Distribution (\%)} \\
\cmidrule(l){3-5}
& & Unchanged & Adapted & Reject \\
\midrule
\multirow{2}{*}{ALFWorld}
& Match ($s^+$) & 46.0 & 53.4 & 0.6 \\
& Edit ($s^-$)  & 37.3 & 56.3 & 6.4 \\
\midrule
\multirow{2}{*}{WebShop}
& Match ($s^+$) & 0.0 & 27.9 & 72.1 \\
& Edit ($s^-$)  & 0.0 & 21.2 & 78.8 \\
\bottomrule
\end{tabular}
\end{minipage}

\caption{\textbf{Analysis of memory reconstruction.}
\textbf{(a)} Ablation on the memory source state $o^{src}$, where RR
and SR denote the per-memory average reject rate and task average
success rate, respectively.
\textbf{(b)} Behavioral shifts under minimal factual edits ($s^-	$).}
\label{tab:reconstruction_analysis}
\end{table*}

%% file: sections/conclusion.tex
\section{Conclusion}
\label{sec:conclusion}

We introduced \textbf{MemHarness}, an RL-driven framework that shifts memory-augmented agents from \emph{verbatim replay} to \emph{state-conditioned reconstruction}. Optimized end-to-end via GRPO, a single unified policy learns to autonomously critique and rewrite historical experiences into state-specific guidance before acting. Experiments on ALFWorld and WebShop show our 7B model significantly outperforms closed-source models and RL baselines, especially in out-of-distribution environments. Analyses reveal this mechanism actively filters mismatched noise and fundamentally improves the agent's intrinsic reasoning, even when memory is unavailable at inference. Future work will explore scaling to larger models and open-ended environments.

%% file: sections/appendix.tex
\newpage
\appendix
\renewcommand{\thesection}{\Alph{section}}
\setcounter{section}{0}

\noindent{\LARGE\textbf{Appendix}\par}\normalsize

\vspace{10pt}
{\large \textbf{Contents}}
\startcontents[appendices]
\printcontents[appendices]{l}{1}{\setcounter{tocdepth}{3}}

\section{Implementation Details}
\label{sec:implementation}

\subsection{Environment and Evaluation Setup}
We evaluate our method on two challenging interactive benchmarks: ALFWorld and WebShop. 
\textbf{ALFWorld}~\cite{shridhar2020alfworld} is an embodied text-based environment containing six sub-task categories (e.g., Pick, Look, Clean). The maximum interaction horizon is set to 50 steps, with a maximum prompt length of 2048 tokens and a generation limit of 512 tokens. 
\textbf{WebShop}~\cite{yao2022web} is an e-commerce website simulation containing over 1.1 million products. The maximum interaction horizon is restricted to 15 steps, with a maximum prompt length of 4096 tokens and a generation limit of 512 tokens.

\textbf{Reward Function Design:} Across both environments, we employ a straightforward outcome-driven reward function combined with a format penalty. The agent receives a sparse outcome reward of $+10$ upon successful task completion and $0$ for failure. To encourage adherence to the prescribed interaction protocol, we add a small formatting bonus, weighted by $0.1$, to the sparse outcome reward. The format score equally weights three criteria: (1) each step contains exactly one valid block followed by one valid \actiontag{action}...\actiontag{/action} block, (2) memory is retrieved through valid \retrievetag{retrieve\_memory}...\retrievetag{/retrieve\_memory} blocks between one and five times per episode, (3) thinking contents are including by one valid \thinktag{think}...\thinktag{/think} block and (4) all generated content is strictly in English.

\subsection{Training Hyperparameters}
\label{sec:hyperparameter}
Our reinforcement learning optimization is based on Group Relative Policy Optimization (GRPO)~\cite{shao2024deepseekmath} and implemented using the veRL~\cite{sheng2024verl} framework. We initialize the policy with instruction-tuned base models Qwen2.5-7B~\cite{qwen2025qwen25technicalreport} and freeze the reference policy $\pi_{ref}$ during training to compute the KL divergence penalty. 

The core hyperparameter settings are consistent across both ALFWorld and WebShop:
\begin{itemize}[leftmargin=1.5em]
    \item \textbf{Rollout \& Sampling:} During the environment interaction phase, we sample a group size of $G=8$ trajectories for each prompt. The rollout generation temperature is set to $1.0$, and the validation temperatue is set to $0.4$. We sample 16 different groups per rollout, utilizing a total of 128 parallel environments.
    \item \textbf{Optimization:} The policy network is updated using the Adam optimizer with a learning rate of $1 \times 10^{-6}$. 
    \item \textbf{GRPO Objectives:} The PPO-style clipping parameter is set to $\epsilon = 0.2$. To prevent the policy from degrading and deviating too far from the base model, the KL divergence penalty coefficient is set to $\beta = 0.01$.
    \item \textbf{Batching:} The training utilizes a mini-batch size of 256 for Alfworld and 64 for Webshop with 1 tensor parallel size. We use vLLM~\cite{kwon2023vllm} for high-throughput rollout generation with a GPU memory utilization ratio of 0.75.
\end{itemize}

\subsection{Memory Bank and Retrieval Setup}
Following prior work, we implement the memory store $\mathcal{B}$ as a vector database using Milvus~\cite{2021milvus} for efficient similarity search. Each memory entry consists of a distilled experience $e_i$ paired with its source interaction state $o_i^{src}$. We use BGE-M3~\cite{chen2024bge} as the embedding model to encode both the retrieval query $q_t$ and the stored entries, and retrieve the top-$k$ ($k=3$) most relevant experiences at each step based on cosine similarity.

\textbf{Memory Initialization.} Inspired by prior work~\cite{wu2025evolver}, the memory bank is not manually curated but is instead populated online through the policy's own interactions. During training, the policy model $\pi_\theta$ distills successful and failed trajectories into concise natural-language principles paired with its source interaction state $o_i^{src}$, which are then inserted into $\mathcal{B}$. To ensure the efficiency of the pipeline, we retain only 50\% of the trajectories generated during the GRPO process for distillation. Specifically, we prioritize a balanced retention of successful and failed trajectories; when such a balance cannot be achieved, the remaining quota is filled by randomly sampling from the leftover trajectories.

\textbf{Deduplication and Pruning.} To prevent the memory bank from growing unboundedly and accumulating redundant entries (e.g., near-identical principles arising from GRPO group sampling), we apply curation at three stages. 
\textit{(1) Write-time deduplication:} before inserting a new memory entry, we perform an embedding-based nearest-neighbor probe; if its maximum cosine similarity to any existing entry exceeds the threshold $\theta_{sim}$ (default $0.85$), the insertion is skipped. In practice, we enable it for WebShop and disable it for ALFWorld. 
\textit{(2) Retrieval-time deduplication:} at inference, we first retrieve a larger candidate pool from bank, then apply greedy embedding-based deduplication before truncating to the top-$k$ results, ensuring the returned memories are not semantically redundant. 
\textit{(3) Capacity and quality maintenance:} we adopt an EvolveR-style Laplace-smoothed utility score $s(p) = (c_{succ}(p)+1)/(c_{use}(p)+2)$, and periodically prune low-utility entries whose score falls below the pruning threshold $\theta_{prune}$ (default $0.3$) once they have been retrieved at least \texttt{min\_uses\_before\_prune} times (default 3).

\subsection{Cold-Start Dataset Construction}
\label{sec:app_cold_start}

We construct a small cold-start dataset to familiarize the policy with the interaction protocol, particularly how to initiate memory retrieval and summarize trajectories into reusable memories. For each benchmark, we collect seed interaction trajectories from the AgentGym dataset~\cite{xi2024agentgym}.

The cold-start dataset contains two subsets, each comprising \texttt{200} examples per benchmark. First, we construct \emph{memory-augmented interaction trajectories}. For each seed trajectory, we randomly sample an intermediate decision points and insert a memory-retrieval turn. GPT-5.1 generates a context-appropriate retrieval query from the task description and interaction prefix at the selected point. The retrieved memory is produced by asking GPT-5.1 to distill the corresponding source trajectory into concise, reusable guidance. The resulting trajectory therefore demonstrates the complete interaction protocol, including issuing a retrieval request, receiving an experience from the memory bank, and continuing environment interaction in the required output format.

Second, we construct 200 \emph{trajectory-to-memory summarization pairs}. Each pair takes a completed AgentGym trajectory as input and uses its GPT-5.1 generated summary as the target, following the structured memory schema described in Appendix~\ref{sec:app_prompts}. These examples teach the policy to convert interaction histories into compact memories containing a situational precondition and a reusable principle.

We combine the two subsets and perform supervised fine-tuning for 2 epochs with a learning rate of $1\times10^{-5}$ and an effective batch size of 2. GPT-5.1 is used only for offline cold-start data construction and is not involved in RL training or evaluation. The purpose of this stage is protocol and format alignment rather than task-skill acquisition; consistent with this role, the cold-start model alone achieves limited task performance, while the actual reconstruction and decision-making capabilities are subsequently learned through task-level reinforcement learning.

\section{Counterfactual Probe Construction}
\label{sec:app_counterfactual}

To evaluate whether MemHarness genuinely relies on fine-grained state comparison rather than superficial pattern matching, we constructed a counterfactual probe dataset comprising 2,000 examples (1,000 for ALFWorld and 1,000 for WebShop). For each instance, we sample a real interaction state ($s^+$) and a genuinely retrieved memory ($m$) that is highly applicable. We then use a strong LLM to minimally edit the state to produce a counterfactual state ($s^-$), rendering $m$ either inapplicable or misleading. 

\subsection{Generation Prompt}
We use GPT-5.1 to generate the counterfactual states. The prompt (Appendix C) enforces that the edits remain minimal and structurally identical to the original observations, preventing the agent from relying on formatting artifacts to reject the memory.

\subsection{Examples of State Edits}

Below we provide concrete examples of the factual edits for both benchmarks. In these examples, the MemHarness agent successfully accepts the memory under the original state ($s^+$) but adaptively rewrites or rejects it when presented with the minimal edit ($s^-$).

\paragraph{ALFWorld Example: Premise Invalidation.} 
In ALFWorld, memories often provide search heuristics based on past failures. The factual edit invalidates the heuristic's premise by revealing the target object.

\begin{itemize}[leftmargin=1.5em]
    \item \textbf{Task:} heat some mug and put it in coffeemachine.
    \item \textbf{Retrieved Memory ($m$):} If you cannot find a mug in nearby cabinets or drawers after opening several of them, shift your search to visible surfaces like countertops or the sinkbasin.
    \item \textbf{Original State ($s^+$):} On the drawer 5, you see nothing.
    \item \textbf{Edited State ($s^-$):} On the drawer 5, you see a mug.
    \item \textbf{Agent Behavior Shift:} In $s^+$, the agent accepts the memory and redirects its search to open surfaces. In $s^-$, the premise (failing to find a mug) is broken; the agent rejects the memory (\emptytag{EMPTY}) and directly picks up the mug.
\end{itemize}

\paragraph{WebShop Example: Constraint Violation.} 
In WebShop, memories often recommend finalizing a purchase when the product matches all instructions. The edit subtly alters a product attribute, violating the user's initial constraint.

\begin{itemize}[leftmargin=1.5em]
    \item \textbf{Task:} Find me machine wash men's t-shirts ... and price lower than 50.00 dollars.
    \item \textbf{Retrieved Memory ($m$):} \textit{After selecting all required attributes ... and confirming the price constraint is satisfied, finish the task by clicking the ``Buy Now'' button.}
    \item \textbf{Original State ($s^+$):} ... ``Price: \$19.99'' [SEP] ``Rating: N.A.'' ... ``Buy Now''
    \item \textbf{Edited State ($s^-$):} ... ``Price: \$59.99'' [SEP] ``Rating: N.A.'' ... ``Buy Now''
    \item \textbf{Agent Behavior Shift:} In $s^+$, the agent follows the memory to click ``Buy Now''. In $s^-$, noticing the price now exceeds the \$50 limit, the agent rejects the memory to prevent a wrong purchase and navigates back to search.
\end{itemize}

\section{Prompt Templates}
\label{sec:app_prompts}

\newcommand{\placeholder}[1]{\texttt{\{#1\}}}

\newtcolorbox{promptbox}[1]{
    enhanced jigsaw, 
    colframe=prompttitlebg,
    colback=promptbodybg,
    coltitle=white,
    fonttitle=\bfseries,
    title=#1,
    arc=2mm,
    boxrule=0.5pt,
    left=3mm, right=3mm, top=3mm, bottom=3mm, 
    toptitle=1.5mm, 
    bottomtitle=1.5mm, 
    lefttitle=3mm, 
    fontupper=\small,
    boxsep=0pt,
    before skip=1em,
    after skip=1em,
    toprule at break=0pt,
    bottomrule at break=0pt,
    pad at break=0mm
}

This section presents the prompt templates used for memory retrieval, state-conditioned memory reconstruction, agent--environment interaction, trajectory summarization, and fallback reasoning when no adapted memory. Text enclosed in braces denotes a runtime placeholder.

\paragraph{Environment feedback Prompt for Agent}
The agent system prompts specify the environment context, interaction history, admissible actions, and the required reasoning--action output format. The templates for ALFWorld and WebShop are presented in Figures~\ref{fig:alfworld-agent-prompt} and~\ref{fig:webshop-agent-prompt}, respectively. The memory retrieval instruction guides the agent to issue a retrieval query when additional experience is needed, while preventing retrieval and environment actions from appearing in the same response.

\input{prompts/system_alfworld}

\input{prompts/system_webshop}

\paragraph{Self-Reasoning Fallback Prompt}
When no retrieved principle is applicable, the fallback prompt instructs the agent to rely on the current observation and its own reasoning instead. The corresponding prompt is shown in Figure~\ref{fig:self_reasoning}.

\input{prompts/self_reasoning}

\paragraph{Contextual Memory Reconstruction Prompt}
The reconstruction prompt adapts a retrieved historical principle to the current task and state, returning \texttt{<EMPTY>} when the principle is not applicable. The full template is presented in Figure~\ref{fig:memory-reconstruction-prompt}.

\input{prompts/reconstruct_memory}

\paragraph{Trajectory Summarization Prompts}
After an episode is completed, the trajectory summarization prompt extracts concise and reusable memories grounded in the interaction trajectory. Its required JSON output schema and extraction constraints are provided in Figure~\ref{fig:trajectory-summarization-prompt}.

\input{prompts/trajectory_summary}

\paragraph{Counterfactual Probe Construction Prompts} 
This section details the prompts used to construct the counterfactual probes for the analysis in Section~\ref{sec:mechanism}. The complete template is shown in Figure~\ref{fig:alfworld-probe-prompt} and Figure~\ref{fig:webshop-probe-prompt}.

\input{prompts/alfworld_probe_prompt}

\input{prompts/webshop_probe_prompt}

\section{Case Studies}
\label{sec:app_case_study}

To intuitively illustrate how MemHarness bridges the gap between historical knowledge and current interaction states, we provide two real reconstruction cases from our evaluation. These examples demonstrate why verbatim replay of past experiences often fails and how our policy actively rewrites memory to provide state-grounded guidance.

\subsection{ALFWorld: Abstracting and Transferring Skills}
In ALFWorld, verbatim replay of the retrieved memory would introduce hallucinated objects (``credit card'' and ``coffee table'') into the context, likely causing the agent to output an invalid action. As shown in Figure~\ref{fig:case-alfworld}, MemHarness instead extracts the underlying procedural skill (placing a held item at the final destination after a prerequisite step) and explicitly grounds it with the current entities (``cup'' and ``sidetable'').
\input{prompts/case_alfworld}

\subsection{WebShop: Grounding Heuristics into Actionable Targets}
In WebShop, the retrieved memory is often an abstract heuristic (e.g., advising the agent to cross-check specific constraints before clicking). A naive replay would simply append this abstract rule to the prompt, offering no concrete operational help. As shown in Figure~\ref{fig:case-webshop}, MemHarness executes this heuristic during the reconstruction phase: it scans the current noisy observation, aligns it with the user's constraints, and directly outputs the exact target ID to act upon.

\input{prompts/case_webshop}

%% file: prompts/system_alfworld.tex
\begin{figure}[t]
\centering
\begin{promptbox}{Alfworld Environment feedback Prompt for Agent}
\small
You are an expert agent operating in the ALFRED Embodied Environment.
Your task is to: \placeholder{task\_description}

Prior to this step, you have already taken \placeholder{step\_count} step(s).
Below are the most recent \placeholder{history\_length} observations and the
corresponding actions you took: \placeholder{action\_history} \\ 

You are now at step \placeholder{current\_step}, and your current observation
is: \placeholder{current\_observation} \\

Your admissible actions in the current situation are: \placeholder{admissible\_actions}. \\

\medskip
Now it is your turn to take an action.
First, reason step by step about the current situation. The reasoning process
must be enclosed within \thinktag{think} and \thinktag{/think} tags.
After finishing the reasoning, choose one admissible action for the current step
and present it within \actiontag{action} and \actiontag{/action} tags. \\

You have access to a memory bank of past experiences from similar tasks. When you are uncertain about how to proceed, prefer retrieving relevant memory for guidance before taking an environment action. To retrieve memory, output exactly one search query wrapped by \retrievetag{retrieve\_memory} and \retrievetag{/retrieve\_memory} tags.
Do not output both a memory retrieval query and an action in the same response.
After the memory is returned, take an environment action in the next response.

\end{promptbox}
\caption{Agent system prompts used in ALFWorld.}
\label{fig:alfworld-agent-prompt}
\end{figure}

%% file: prompts/system_webshop.tex
\begin{figure}[t]
\centering
\begin{promptbox}{Webshop Environment feedback Prompt for Agent}
\small
You are an expert autonomous agent operating in the WebShop e-commerce
environment.

Your task is to: \placeholder{task\_description}.

Prior to this step, you have already taken \placeholder{step\_count} step(s).
Below are the most recent \placeholder{history\_length} observations and the
corresponding actions you took: \placeholder{action\_history} \\

You are now at step \placeholder{current\_step}, and your current observation
is: \placeholder{current\_observation}.\\

Your admissible actions in the current situation are: \texttt{[}\;\placeholder{available\_actions}\;\texttt{]}. \\

Now it is your turn to take one action for the current step.
First, reason step by step about the current situation and determine which
admissible action best advances the shopping goal. The reasoning process must
be enclosed within \thinktag{think} and \thinktag{/think} tags.
After finishing the reasoning, choose one admissible action and present it
within \actiontag{action} and \actiontag{/action} tags. \\

You have access to a memory bank of past experiences from similar tasks. When you are uncertain about how to proceed, prefer retrieving relevant memory for guidance before taking an environment action. To retrieve memory, output exactly one search query wrapped by \retrievetag{retrieve\_memory} and \retrievetag{/retrieve\_memory} tags.
Do not output both a memory retrieval query and an action in the same response.
After the memory is returned, take an environment action in the next response.

\end{promptbox}
\caption{Agent system prompts used in WebShop, with and without recent
interaction history.}
\label{fig:webshop-agent-prompt}
\end{figure}

%% file: prompts/self_reasoning.tex
\begin{figure}[!ht]
\centering
\begin{promptbox}{Self-Reasoning Fallback Instruction}
\small

No validated memory principle applies; rely on observation and reasoning.

\end{promptbox}
\caption{Prompt used to self-reasoning fallback}
\label{fig:self_reasoning}
\end{figure}

%% file: prompts/reconstruct_memory.tex
\begin{figure*}[!ht]
\centering
\begin{promptbox}{State-Conditioned Memory Reconstruction Prompt}
\small

\textbf{System Prompt} \\[2pt]
You adapt a retrieved memory principle into concise, reusable guidance for the
current situation and initial task, or output exactly \texttt{<EMPTY>} if the
principle does not apply.

Do not write chain-of-thought, first-person reasoning, or a step-by-step action
plan.

\medskip
\textbf{User Message Template} \\[2pt]
\textbf{Initial task:} \placeholder{task} \\
\medskip
\textbf{Current situation (state):} \placeholder{s\_curr} \\
\medskip
\textbf{Retrieved historical state:} \placeholder{s\_old} \\
\medskip
\textbf{Historical principle (memory):} \placeholder{p\_old} \\
\medskip
\textbf{Output Rules:} 
\begin{itemize}
    \setlength{\itemsep}{1pt}
    \item Output exactly one short adapted principle, or exactly
          \emptytag{EMPTY}.
    \item Write in an imperative or neutral guidance style. \\
    \item Do not use first person (e.g., ``I,'' ``my,'' or ``we'') or phrases
          such as ``therefore,'' ``I need to,'' or ``since''.
    \item Do not explain the reasoning or describe multiple next steps.
\end{itemize}

\end{promptbox}
\caption{Prompt used to reconstruct a retrieved memory principle according to
the current environment state.}
\label{fig:memory-reconstruction-prompt}
\end{figure*}

%% file: prompts/trajectory_summary.tex
\begin{figure*}[t]
\centering
\begin{promptbox}{Trajectory Summarization Prompt}
\small

\textbf{System Prompt} \\[2pt]
You are a JSON-only memory extractor.

Read a completed agent trajectory and write reusable advice for future similar states.

\medskip
\textbf{Critical Rules:}
\begin{itemize}
    \setlength{\itemsep}{0pt}
    \item Return exactly one JSON object and nothing else.
    \item Do not continue the trajectory.
    \item Do not write thoughts, actions, markdown, XML tags, or explanations.
    \item Do not copy raw \thinktag{think}, \actiontag{action}, or \retrievetag{retrieve\_memory} text.
    \item Each memory must be grounded in the trajectory and useful later.
\end{itemize}

\medskip
\hrule
\medskip

\textbf{User Message Template} \\[2pt]
Benchmark: ``\placeholder{task\_name}''.

Task: extract at most \placeholder{num\_memories} reusable memories from the completed trajectory below.

Return only this JSON shape:
{\ttfamily
\{\\
\hspace*{1em}``memories'': [\\
\hspace*{2em}\{\\
\hspace*{3em}``situation'': ``one short sentence ...'',\\
\hspace*{3em}``memory'': ``one short sentence ...''\\
\hspace*{2em}\}\\
\hspace*{1em}]\}
}

\textbf{Field Rules:}
\begin{itemize}
    \setlength{\itemsep}{0pt}
    \item \texttt{situation}: generalized state or precondition, not a full observation dump.
    \item \texttt{memory}: reusable advice, not a recap and not a next action command.
    \item If no useful memory exists, return \texttt{\{``memories'': []\}}.
\end{itemize}
\medskip
\textbf{Trajectory:} \\
\placeholder{trajectory\_text}

\end{promptbox}
\caption{Prompt used for offline trajectory summarization. The model distills raw interaction experiences into a structured JSON format containing situational preconditions and reusable guidance.}
\label{fig:trajectory-summarization-prompt}
\end{figure*}

%% file: prompts/alfworld_probe_prompt.tex
\begin{figure*}[!t]
\centering
\begin{promptbox}{ALFWorld Counterfactual Probe Generation Prompt}
\small
\textbf{[System Instructions]} \\
You create ONE counterfactual mismatched observation for a memory probe.

You are given a matched current observation (\texttt{s\_plus}) where a retrieved memory applies. Produce \texttt{s\_minus}: \texttt{s\_plus} with ONE minimal factual change so the memory NO LONGER applies.

Rules: \\
- Keep the same observation style as \texttt{s\_plus} (AlfWorld), including entity names, ids, and length. \\
- Change only one applicability-related fact (see memory preconditions / principle). \\
- Do NOT produce unnatural or gibberish text. \\
- Output valid JSON only (no markdown fences): \\
\texttt{\{``s\_minus'': ``...'', ``flip'': ``one-line description of the single fact you changed''\}}

\medskip
\textbf{[User Input]} \\
Initial task (fixed): \placeholder{task} 

Retrieved historical state s\_old (fixed; NOT the current observation): \placeholder{s\_old} 

Memory principle p\_old: \placeholder{p\_old} 

Matched current observation s\_plus (fixed; do NOT rewrite): \placeholder{s\_plus} 

Memory metadata (for applicability): \placeholder{metadata\_json} 

Return JSON with exactly: s\_minus, flip.
\end{promptbox}
\caption{Prompt used to generate counterfactual mismatched observations for ALFWorld.}
\label{fig:alfworld-probe-prompt}
\end{figure*}

%% file: prompts/webshop_probe_prompt.tex
\begin{figure*}[t]
\centering
\begin{promptbox}{WebShop Counterfactual Probe Generation Prompt}
\small
\textbf{[System Instructions]} \\
You create ONE counterfactual mismatched WebShop observation for a memory probe.

You are given a matched current observation (\texttt{s\_plus}) where a retrieved memory applies. Produce \texttt{s\_minus}: \texttt{s\_plus} with ONE minimal factual change so the memory NO LONGER applies.

WebShop format rules: \\
- \texttt{s\_minus} MUST look like a WebShop formatted observation: 'segment' [SEP] ``segment'' [SEP] ... \\
- Every segment MUST be wrapped in single quotes. \\
- Preserve [SEP] separators exactly as in \texttt{s\_plus}. \\
- Change only ONE segment/value (or remove/add one segment) so the memory no longer applies. \\
- Do NOT output plain prose (e.g. ``No Search'', ``nothing available'') without quoted [SEP] segments. \\
- Keep overall length and structure similar to \texttt{s\_plus}. \\

Output valid JSON only (no markdown fences):
\texttt{\{``s\_minus'': ``...'', ``flip'': ``one-line description of the single fact you changed''\}}

\medskip
\textbf{[User Input]} \\
Initial task (fixed): \placeholder{task} 

Retrieved historical state s\_old (fixed; NOT the current observation):
\placeholder{s\_old} 

Memory principle p\_old: \placeholder{p\_old} 

Matched current observation s\_plus (fixed; do NOT rewrite): \placeholder{s\_plus}

Memory metadata (for applicability): \placeholder{metadata\_json}

Return JSON with exactly: s\_minus, flip.
\end{promptbox}
\caption{Prompt used to generate counterfactual mismatched observations for WebShop.}
\label{fig:webshop-probe-prompt}
\end{figure*}

%% file: prompts/case_alfworld.tex
\begin{figure}[t]
\centering
\begin{promptbox}{Case Study: ALFWorld Skill Transfer}
\small
\textbf{Intial Task:} \\ 
cool some cup and put it in sidetable 

\medskip
\textbf{Current State ($s_t$):} \\
You cool the cup 2 using the fridge 1. 

\medskip
\textbf{Historical State ($s_{src}$):} \\
Holding a creditcard in inventory; at the coffeetable where two credit cards need to be placed on it; the task is to put two credit cards on the coffeetable. 

\medskip
\textbf{Retrieved Memory ($m$):} \\
\textit{Once you have acquired and prepared the target object (e.g., a credit card), proceed directly to the final receptacle (e.g., the coffee table) and place it there to complete the goal.}

\vspace{0.5em}
\hrule
\vspace{0.5em}

\textbf{MemHarness Reconstruction ($g_t$):} \\
If you are holding a target object (here, a cup) and you have already used the fridge to cool it, the next step is to place it at the destination location (here, the sidetable). The fridge is not required in this step; it was only needed to cool the cup. The sidetable is the goal location where the cup should be placed.
\end{promptbox}
\caption{ALFWorld case demonstrating how MemHarness prevents hallucination by rewriting a mismatched historical memory into applicable, grounded guidance.}
\label{fig:case-alfworld}
\end{figure}







%% file: prompts/case_webshop.tex
\begin{figure}[!ht]
\centering
\begin{promptbox}{Case Study: WebShop Heuristic Grounding}
\small
\textbf{Intial Task:} \\
Find me day comfort, anti slip, non slip women's oxfords with high heel, closed toe, ankle strap, memory foam, rubber sole with color: black, and size: 6

\medskip
\textbf{Current State ($s_t$):} \\
\texttt{``Back to Search'' [SEP] ``Page 1 (Total results: 50)'' [SEP] ``Next >'' [SEP] ``B06W51MMKY'' [SEP] ``Alegria Women's Kourtney'' [SEP] ``\$62.99 to \$129.99'' [SEP] ... [Other 9 products omitted for brevity] ...}

\medskip
\textbf{Historical State ($s_{src}$):} \\
\texttt{On a WebShop search results page with many clothing items listed only by ID, color, size, and price ranges, with no visible product details yet.}

\medskip
\textbf{Retrieved Memory ($m$):} \\
\textit{When facing a search results page with mixed items and highly detailed constraints (e.g., fit, fabric, color, size), carefully cross-reference the visible titles and attributes against all instructions before clicking a specific product ID.} 

\vspace{0.5em}
\hrule
\vspace{0.5em}

\textbf{MemHarness Reconstruction ($g_t$):} \\
B06W51MMKY: Alegria Women's Kourtney - seems to match the criteria of day comfort, anti slip, non slip women's oxfords with high heel, closed toe, ankle strap, memory foam, rubber sole with color: black, and size: 6.
\end{promptbox}
\caption{WebShop case demonstrating how MemHarness translates an abstract heuristic rule into a concrete, actionable target based on the current observation.}
\label{fig:case-webshop}
\end{figure}





